\documentclass{article}

\PassOptionsToPackage{numbers}{natbib}
\usepackage[preprint]{neurips_2026}

\usepackage[export]{adjustbox}
\usepackage[inline, shortlabels]{enumitem}
\usepackage[T1]{fontenc}
\usepackage{hyperref}
\usepackage{microtype}
\usepackage{xurl}

\usepackage{amsmath, amsfonts}
\usepackage{nicefrac}

\usepackage{graphicx}
\usepackage{booktabs}
\usepackage{multirow}
\usepackage{array}       
\usepackage{xcolor}
\usepackage{colortbl}
\makeatletter
\@ifundefined{insert@pcolumn}{\let\insert@pcolumn\insert@column}{}
\makeatother

\usepackage{tikz}
\usepackage{pgfplots}
\pgfplotsset{compat=1.18}
\usetikzlibrary{shapes.geometric, arrows.meta, positioning, fit, backgrounds}

\title{GxP-Agent: Process-DAG Topology for Reliable Clinical Trial Programming with LLM Agents}

\author{%
  Jaime Yan \\
  Data Science Researcher \\
  Harrisburg University of Science and Technology \\
  ORCID: \href{https://orcid.org/0009-0007-1786-7259}{0009-0007-1786-7259}
}

\begin{document}

\maketitle

\begin{abstract}
Clinical trial programming---transforming study protocols into analysis-ready datasets under CDISC standards---is a bottleneck in regulatory submissions, yet LLM-based code generation fails catastrophically on this task: across 11 single-shot attempts with five frontier models, none produces a valid subject-level analysis dataset.
We introduce GxP-Agent, a multi-agent system that encodes regulatory process ordering as a directed acyclic graph (DAG), decomposing monolithic dataset generation into 15 domain-specific nodes executed by worker agents with pharmaverse skill context, validation gates, and conditional retry.
On CDISC-Bench, a new execution-based benchmark built from the FDA pilot submission CDISCPilot01 (254 subjects, 49 ground-truth ADSL variables), GxP-Agent with Claude Sonnet~4.6 achieves 100\% structural match (49/49 variables, 254 correct records) across three independent runs, compared to 59.2\% for the best retrieval-augmented baseline and 0\% for all single-agent and flat multi-agent approaches.
The DAG topology also enables weaker models: GPT-4.1 achieves 59.2\% mean structural match under the same DAG, where it scores 0\% under every other architecture.
The approach generalizes to ADAE (adverse events; 9-node branching DAG, 55 variables, 1{,}191 records), achieving 100\% structural match on the first attempt.
These results demonstrate that encoding domain process knowledge as graph topology---rather than relying on LLM reasoning alone---is a key enabler for reliable, GxP-compliant clinical trial programming.
\end{abstract}

\section{Introduction}

Clinical trials generate the evidence base for approximately 80\% of FDA drug approvals~\citep{fda2023cder}.
Each submission requires transforming raw clinical data into analysis-ready datasets following CDISC ADaM standards, then producing Tables, Listings, and Figures (TLFs) per the Statistical Analysis Plan.
This programming is labor-intensive, requiring weeks of effort by specialized statistical programmers under strict GxP compliance requirements~\citep{ich2016e6r2}.

Can frontier LLMs automate this task?
Our experiments reveal a sobering answer: across five models (Claude Sonnet~4.6, Claude Haiku~4.5, GPT-4.1, GPT-4o, Gemini~2.5~Pro), single-shot generation of even ADSL (the subject-level analysis dataset) achieves a \textbf{0\% pass rate} on 11 independent attempts.
Manual inspection of these 11 failures reveals recurring error categories: variable name hallucination ($\sim$40\% of errors), function signature hallucination ($\sim$30\%), incorrect file references ($\sim$15\%), and logic errors ($\sim$15\%)---reflecting the difficulty of coordinating dozens of interdependent derivations within a single generation.

We propose encoding the regulatory-mandated process ordering as a \textbf{directed acyclic graph (DAG)}, assigning each derivation step to a specialized worker agent.
Clinical trial programming is not a creative task where open-ended reasoning helps, but a \textbf{compliance task} where correct ordering, complete coverage, and verifiable intermediate outputs are paramount.
This setting provides a natural testbed for a fundamental question in LLM agent design: when tasks have known dependency structure, does encoding that structure as agent topology outperform letting LLMs plan their own decomposition?

\noindent\textbf{Contributions.}
\begin{itemize}[leftmargin=1.5em, itemsep=2pt]
    \item \textbf{GxP-Agent}, a multi-agent system compiling regulatory process specifications into executable DAGs with node-type-specific prompts, validation gates, schema introspection, and conditional retry (\S\ref{sec:system-design}).
    \item \textbf{CDISC-Bench}, an execution-based benchmark from the FDA CDISCPilot01 pilot submission with ground-truth ADSL (254 records, 49 variables) and five-level evaluation (\S\ref{sec:benchmark}).
    \item \textbf{Empirical evidence} that DAG topology is decisive: 100\% structural match with Claude (1 Opus + 3 Sonnet runs, 0\% variance), 59.2\% with GPT-4.1, vs.\ 0\% for all non-DAG architectures except Keyword-RAG+Sonnet (59.2\%) (\S\ref{sec:experiments}).
    \item \textbf{Analysis} showing DAG decomposition disproportionately benefits weaker models by isolating failures to individual nodes (\S\ref{sec:discussion}).
\end{itemize}

\section{Background}

\textbf{CDISC standards.}
CDISC defines two data models for regulatory submissions~\citep{cdisc2021adam}: \textbf{SDTM} standardizes raw clinical data into domains (DM, AE, EX, LB, etc.); \textbf{ADaM} defines analysis-ready datasets derived from SDTM, with ADSL (Subject-Level Analysis Dataset) as the foundational dataset containing one record per subject.
ADSL generation requires coordinating derivations across multiple SDTM domains (treatment dates from EX, disposition from DS, vital signs from VS, questionnaire scores from QS), each following specific business rules and CDISC controlled terminology.
Our benchmark ADSL contains 49 variables derived from 8 SDTM domains---requiring coordination across dozens of interdependent derivations, a scale that exceeds current LLM single-context reliability (\S\ref{sec:experiments}).

\textbf{Pharmaverse.}
The pharmaverse is a collection of open-source R packages for clinical trial reporting~\citep{pharmaverse2024}: \textbf{admiral}~\citep{admiral2024} (ADaM derivation functions), \textbf{metacore}~\citep{metacore2024} (metadata specifications), \textbf{metatools}~\citep{metatools2024} (metadata utilities), and \textbf{xportr}~\citep{xportr2024} (XPT export for regulatory submission).

\textbf{LLM agents and graph orchestration.}
Recent work explores LLM agents for software engineering~\citep{jimenez2024swebench} and multi-agent collaboration~\citep{wu2023autogen,hong2024metagpt}.
Graph frameworks such as LangGraph~\citep{langgraph2024} enable agent workflows as state machines with conditional edges and retry logic.
However, existing multi-agent systems rely on LLM reasoning for task decomposition.
Our approach differs: decomposition and ordering are \textbf{predetermined by domain knowledge}, not discovered by the LLM at runtime, eliminating planning failures and providing structural guarantees.

\section{System Design}
\label{sec:system-design}

\subsection{Architecture Overview}

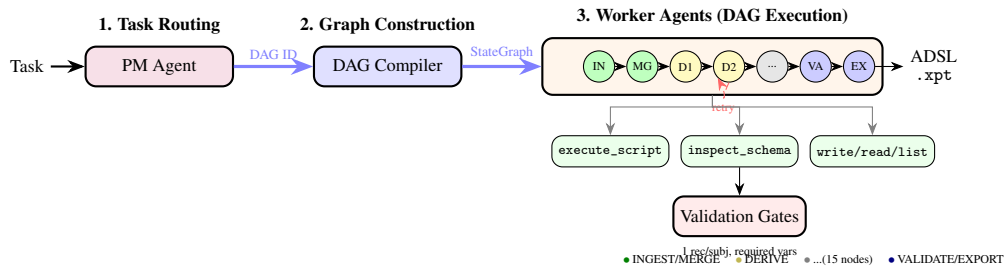
\begin{figure}[t]
\centering
\begin{tikzpicture}[
    scale=0.75, every node/.style={transform shape},
    node distance=0.4cm and 0.8cm,
    >=Stealth,
    box/.style={draw, rounded corners, minimum height=0.7cm, minimum width=2.0cm, align=center, font=\small},
    component/.style={box, thick},
    tool/.style={draw, rounded corners, fill=green!8, font=\scriptsize, minimum width=1.6cm, minimum height=0.55cm},
    dagnode/.style={draw, circle, minimum size=0.55cm, font=\scriptsize, inner sep=1pt},
    lbl/.style={font=\footnotesize\bfseries, align=center},
    arr/.style={->, thick},
]

\node[component, fill=purple!12, minimum width=2.6cm] (pm) {PM Agent};
\node[lbl, above=0.1cm of pm] {1. Task Routing};

\node[component, fill=blue!12, minimum width=2.6cm, right=1.4cm of pm] (compiler) {DAG Compiler};
\node[lbl, above=0.1cm of compiler] {2. Graph Construction};

\node[component, fill=orange!8, minimum width=6.0cm, minimum height=1.0cm, right=1.4cm of compiler] (workers) {};
\node[lbl, above=0.1cm of workers] {3. Worker Agents (DAG Execution)};

\node[dagnode, fill=green!25] at ([xshift=-2.0cm]workers.center) (n1) {\tiny IN};
\node[dagnode, fill=green!25, right=0.2cm of n1] (n2) {\tiny MG};
\node[dagnode, fill=yellow!30, right=0.2cm of n2] (n3) {\tiny D1};
\node[dagnode, fill=yellow!30, right=0.2cm of n3] (n4) {\tiny D2};
\node[dagnode, fill=gray!20, right=0.2cm of n4] (n5) {\tiny ...};
\node[dagnode, fill=blue!20, right=0.2cm of n5] (n6) {\tiny VA};
\node[dagnode, fill=blue!20, right=0.2cm of n6] (n7) {\tiny EX};

\foreach \i/\j in {n1/n2, n2/n3, n3/n4, n4/n5, n5/n6, n6/n7} {
    \draw[->, thin] (\i) -- (\j);
}

\draw[->, thin, dashed, red!60] (n4.south) .. controls +(-0.2,-0.4) and +(0.2,-0.4) .. (n4.south west)
    node[below, font=\tiny, pos=0.5] {retry};

\node[right=0.5cm of n7, font=\small, align=center] (output) {ADSL\\[-1pt]\texttt{.xpt}};
\draw[arr, thin] (n7) -- (output);

\node[left=0.6cm of pm, font=\small, align=center] (input) {Task};
\draw[arr] (input) -- (pm);
\draw[arr, very thick, blue!50] (pm) -- node[above, font=\scriptsize] {DAG ID} (compiler);
\draw[arr, very thick, blue!50] (compiler) -- node[above, font=\scriptsize] {StateGraph} (workers);

\node[tool, below=0.7cm of workers, xshift=-1.8cm] (t1) {\texttt{execute\_script}};
\node[tool, right=0.2cm of t1] (t2) {\texttt{inspect\_schema}};
\node[tool, right=0.2cm of t2] (t3) {\texttt{write/read/list}};

\draw[arr, thin, gray] (workers.south) ++(0,0) -- ++(0,-0.2) -| (t1.north);
\draw[arr, thin, gray] (workers.south) ++(0,0) -- ++(0,-0.2) -| (t2.north);
\draw[arr, thin, gray] (workers.south) ++(0,0) -- ++(0,-0.2) -| (t3.north);

\node[component, fill=red!8, minimum width=2.2cm, below=0.5cm of t2] (gates) {Validation Gates};
\node[font=\tiny, below=0.05cm of gates] {1 rec/subj, required vars};
\draw[arr, thin] (t2.south) -- (gates.north);

\node[font=\tiny, anchor=north west] at ([xshift=-1cm, yshift=-0.2cm]gates.south west) {%
    \textcolor{green!50!black}{$\bullet$}\,INGEST/MERGE \quad
    \textcolor{yellow!70!black}{$\bullet$}\,DERIVE \quad
    \textcolor{gray}{$\bullet$}\,...(15 nodes) \quad
    \textcolor{blue!50!black}{$\bullet$}\,VALIDATE/EXPORT%
};

\end{tikzpicture}
\caption{GxP-Agent architecture. The PM Agent selects a process DAG from a registry of 11 DAGs. The DAG Compiler produces a LangGraph StateGraph with conditional retry edges. Worker Agents execute each node with six tools (R execution, schema introspection, file management) and schema context injection. Validation Gates enforce structural assertions after critical nodes.}
\label{fig:architecture}
\end{figure}

GxP-Agent consists of four components (Figure~\ref{fig:architecture}):

\textbf{(1) Project Manager Agent} receives a task specification and selects the appropriate process DAG from a registry of pre-defined DAGs (currently containing 11 templates, of which ADSL is evaluated here; see \S\ref{sec:limitations}), then compiles it into an executable graph.

\textbf{(2) DAG Compiler} translates the selected DAG into a LangGraph \texttt{StateGraph}.
Each process step becomes an agent invocation with a system prompt constructed from the node's type, description, and pinned skills.
Conditional routing directs success to the next topological node, failure to retry (up to 2 attempts with the error traceback appended to the original prompt), and exhausted retries to skip.
Before each node executes, the compiler introspects the workspace (\texttt{.rds}/\texttt{.xpt} files, column schemas) and injects this context into the worker prompt.

\textbf{(3) Worker Agents} execute each node as a LangGraph \texttt{create\_react\_agent} with tools for R code execution, data schema inspection, and file management.
After each node, a post-execution check validates that the output \texttt{.rds} file contains expected columns.

\textbf{(4) Validation Gates} run domain-specific R assertions after critical nodes.
These check three assertion classes: (i)~record-level (one record per subject, expected row count), (ii)~variable-level (required ADaM variables present with correct types), and (iii)~business-rule-level (e.g., SAFFL=\texttt{``Y''} implies non-missing TRTSDT).
The ADSL DAG uses 12 such assertions derived from the ADaM Implementation Guide.

\subsection{ADSL DAG Design}

The ADSL process DAG decomposes subject-level dataset generation into 15 nodes in topological order:

\begin{small}
\begin{verbatim}
ingest -> merge_dm -> derive_treatment_vars ->
derive_trt_dates -> derive_disposition ->
derive_completion -> derive_duration ->
derive_flags -> derive_demographics ->
derive_site -> derive_baselines ->
derive_study_dates -> validate ->
apply_metadata -> export
\end{verbatim}
\end{small}

Each node loads the previous node's \texttt{.rds} output, adds its derived columns, and saves the updated file.
This incremental build ensures each worker operates on 2--5 new variables rather than all 49 simultaneously.
The ADSL DAG is linear (purely sequential); other ADaM datasets (e.g., ADAE, which derives from both ADSL and AE domains) require branching topologies with parallel derivation paths. We use the DAG formalism to generalize across both patterns.
Node-type-specific prompts are critical: DERIVE\_DATE nodes receive ``use \texttt{as.Date()}, NOT admiral date functions''; INGEST nodes save only the primary data frame; MERGE nodes load \texttt{.xpt} files directly.

\subsection{Design Rationale}

The DAG topology provides four structural guarantees that LLM-based planning cannot:
\textbf{(1) Completeness}---every required derivation has a dedicated node;
\textbf{(2) Ordering}---topological sort ensures data dependencies are respected;
\textbf{(3) Isolation}---a failure in one node does not corrupt the workspace;
\textbf{(4) Debuggability}---each node produces a traceable artifact for GxP compliance audit trails.

\section{CDISC-Bench}
\label{sec:benchmark}

\subsection{Benchmark Construction}

CDISC-Bench is an execution-based benchmark for evaluating LLM systems on clinical trial programming tasks.
It is built from CDISCPilot01, the FDA's publicly available CDISC pilot submission containing data from a clinical trial of 306 enrolled subjects (254 in the intent-to-treat population) across 3 treatment arms~\citep{cdiscpilot01}.

\textbf{Source data.} 22 SDTM domain \texttt{.xpt} files (DM, AE, EX, DS, LB, VS, SC, QS, SV, MH, and others) containing tabulated clinical data.

\textbf{Ground truth.} Production-quality ADaM datasets (ADSL with 254 records and 49 variables, ADAE with 1,191 records, ADLBC with 7,778 records) included in the pilot submission.

\textbf{Specifications.} The ADaM Specifications document (converted to markdown) defines the derivation logic for each variable, including source domains, business rules, and controlled terminology.

\subsection{Evaluation Framework}

We define a five-level evaluation framework with increasing stringency (Table~\ref{tab:eval-levels}).

\begin{table}[t]
    \centering
    \caption{CDISC-Bench five-level evaluation framework.}
    \label{tab:eval-levels}
    \begin{tabular}{@{}llp{7.5cm}@{}}
        \toprule
        Level & Name & What It Measures \\
        \midrule
        L1 & Routing & Does the system select the correct process DAG and skills? \\
        L2 & Code Quality & Does generated R code use correct packages, functions, and idioms? \\
        L3 & Execution & Does the code execute without errors in R~4.4.1? \\
        L4 & Structural Match & Does the output contain the correct records and variables? \\
        L5 & Value Match & Do computed values match ground truth? \\
        \bottomrule
    \end{tabular}
\end{table}

This paper focuses on L3 (execution) and L4 (structural match), which together determine whether the generated dataset is structurally correct.

\textbf{Structural match} is defined as the fraction of the 49 ground-truth ADSL variable \emph{names} present in the output dataset (measuring variable coverage, not value correctness):
\begin{equation}
    \text{Struct\%} = \frac{|\text{output\_vars} \cap \text{GT\_vars}|}{|\text{GT\_vars}|} \times 100
\end{equation}

A 100\% structural match means all expected variable names are present; it does not guarantee that all computed values match ground truth.
We therefore also verify L5 spot checks on specific values: placebo arm count (expected: 86), mean age in placebo arm (expected: 75.21), and ITT flag count (expected: 254).

\section{Experiments}
\label{sec:experiments}

\subsection{Experimental Setup}

\textbf{Models.} Five models for single-shot baselines (Claude Sonnet~4.6, Claude Haiku~4.5~\citep{anthropic2024claude3}, GPT-4.1, GPT-4o~\citep{openai2023gpt4}, Gemini~2.5~Pro~\citep{google2024gemini}) and five for DAG evaluation (Claude Opus~4.6, Claude Sonnet~4.6~\citep{anthropic2024claude3}, GPT-4.1, GPT-4o~\citep{openai2023gpt4}, Gemini~2.5~Pro~\citep{google2024gemini}), all via API.
\textbf{Temperature:} 0 for Claude and OpenAI models; 0.2 for Gemini (minimum supported value).

\textbf{Architectures:}
\textbf{SingleAgent}---one LLM call generates the complete ADSL R script;
\textbf{FlatMulti}---multiple LLM calls (one per spec section), concatenated;
\textbf{Keyword-RAG}---keyword-matched pharmaverse documentation injected into the prompt;
\textbf{Embedding-RAG}---ChromaDB with all-MiniLM-L6-v2 embeddings retrieves the 15 most similar pharmaverse function docs by cosine similarity;
\textbf{DAG (ours)}---GxP-Agent with the 15-node ADSL DAG.

\textbf{Environment.} R~4.4.1 with admiral~1.4.1, metacore~0.2.1, metatools~0.2.0, xportr~0.5.0, isolated R subprocesses.
DAG experiments: Claude Sonnet (3 runs), GPT-4.1 (3 runs), and GPT-4o (3 runs); other configurations run once due to API cost.

\textbf{Compute.} API-based only (no local GPU). Wall-clock: 65s (Keyword-RAG) to 662s (DAG+Sonnet). Total project cost: $\sim$\$80 USD across $\sim$50 runs (including development iterations). Per-run estimates: SingleAgent $\sim$\$0.05, Keyword-RAG $\sim$\$0.05, DAG+Sonnet $\sim$\$0.25, DAG+Opus $\sim$\$1.35 (15 sequential LLM calls with $\sim$5K input tokens per node). R execution: single CPU core, $<$1 min/node.

\textbf{Implementation note (important).} DAG experiments use the same topology, prompts, retry logic, and validation as the full GxP-Agent system, but invoke LLMs directly per node rather than through the LangGraph tool-calling agent loop. Schema context is pre-injected. This setup performs a controlled evaluation of the DAG architecture, isolating the core effects of topological decomposition from potential confounding factors introduced by agent-loop overhead.

\textbf{L1--L2 validation.}
L1 routing: 87.5\% (GxP DAG router) and 93.8\% (GPT-4.1 LLM router).
L2 code quality scores: 0.076 (Gemini Flash) to 0.519 (GPT-4o), Claude Sonnet 0.492.
The remainder focuses on L3--L4 end-to-end results.

\subsection{Main Results}

Table~\ref{tab:main-results} presents the architecture-by-model comparison.

\begin{table}[t]
    \centering
    \caption{Architecture $\times$ Model on ADSL (254 records, 49 variables). Struct\% = variable name coverage.}
    \label{tab:main-results}
    \small
    \begin{tabular}{@{}llcrrrrr@{}}
        \toprule
        Architecture & Model & Exec & Rec. & Vars & Struct\% & Time & Runs \\
        \midrule
        SingleAgent & Sonnet 4.6 & FAIL & 0 & 0/49 & 0.0 & 119s & 1 \\
        SingleAgent & GPT-4.1 & FAIL & 0 & 0/49 & 0.0 & 95s & 1 \\
        FlatMulti & Sonnet 4.6 & FAIL & 0 & 0/49 & 0.0 & 279s & 1 \\
        FlatMulti & GPT-4.1 & FAIL & 0 & 0/49 & 0.0 & 59s & 1 \\
        Keyword-RAG & Sonnet 4.6 & OK & 254 & 29/49 & 59.2 & 65s & 1 \\
        Keyword-RAG & GPT-4.1 & FAIL & 0 & 0/49 & 0.0 & 64s & 1 \\
        Embedding-RAG & Sonnet 4.6 & FAIL & 0 & 0/49 & 0.0 & 37s & 1 \\
        \midrule
        \textbf{DAG} & \textbf{Opus 4.6} & \textbf{OK} & \textbf{254} & \textbf{49/49} & \textbf{100.0} & \textbf{654s} & \textbf{1} \\
        \textbf{DAG} & \textbf{Sonnet 4.6} & \textbf{OK} & \textbf{254} & \textbf{49/49} & \textbf{100.0} & \textbf{662s} & \textbf{3} \\
        DAG & GPT-4.1 & OK & 254 & 29/49 & 59.2 & 369s & 3 \\
        DAG & GPT-4o & OK & 254 & 16/49 & 33.3 & 280s & 3 \\
        DAG & Gemini 2.5 Pro & FAIL & 0 & 0/49 & 0.0 & 581s & 1 \\
        \bottomrule
    \end{tabular}
\end{table}

\textbf{(1) Single-shot fails universally.}
Across 16 single-shot attempts (11 base + retry/metadata variants) with 5 models, none produces a valid ADSL.

\textbf{(2) DAG topology is decisive.}
Claude Sonnet scores 0\% under SingleAgent, FlatMulti, and Embedding-RAG but 100\% under DAG---same model, different architecture.
Embedding-RAG retrieves 15 semantically relevant skill documents but still produces 0\% structural match, confirming that retrieval alone cannot substitute for structural decomposition.
Fisher's exact test: DAG vs.\ non-DAG, $p = 0.011$.

\textbf{(3) DAG enables weaker models.}
GPT-4.1 scores 0\% under all non-DAG architectures but 59.2\% under DAG.
Cliff's $d = 0.267$ (Claude vs.\ GPT-4.1 node completion).

\textbf{(4) Perfect Claude reproducibility.}
All 4 DAG runs (1 Opus, 3 Sonnet) achieve 100\% with correct value spot-checks. Retry handles residual nondeterminism (6/45 Sonnet node-executions required a second attempt).

\subsection{Per-Node Analysis}

Figure~\ref{fig:heatmap} shows per-node success rates (full data in Appendix Table~\ref{tab:per-node}).


\begin{figure}[t]
\centering
\setlength{\tabcolsep}{4pt}
\renewcommand{\arraystretch}{1.1}

\newcommand{\cellgreen}[1]{\cellcolor{green!30}#1}
\newcommand{\cellyellow}[1]{\cellcolor{yellow!40}#1}
\newcommand{\cellorange}[1]{\cellcolor{orange!40}#1}
\newcommand{\cellred}[1]{\cellcolor{red!30}#1}

\small
\begin{tabular}{@{}l|ccccc@{}}
    \toprule
    \textbf{Node} & \rotatebox{60}{\textbf{Opus}} & \rotatebox{60}{\textbf{Sonnet}} & \rotatebox{60}{\textbf{GPT-4.1}} & \rotatebox{60}{\textbf{GPT-4o}} & \rotatebox{60}{\textbf{Gemini}} \\
    & \scriptsize(1) & \scriptsize(3) & \scriptsize(3) & \scriptsize(3) & \scriptsize(1) \\
    \midrule
    ingest              & \cellgreen{100} & \cellgreen{100} & \cellgreen{100} & \cellgreen{100} & \cellgreen{100} \\
    merge\_dm           & \cellgreen{100} & \cellgreen{100} & \cellgreen{100} & \cellgreen{100} & \cellgreen{100} \\
    derive\_trt\_vars    & \cellgreen{100} & \cellgreen{100} & \cellgreen{100} & \cellgreen{100} & \cellgreen{100} \\
    derive\_trt\_dates   & \cellgreen{100} & \cellgreen{100} & \cellyellow{67}  & \cellorange{33} & \cellgreen{100} \\
    derive\_disposition  & \cellgreen{100} & \cellgreen{100} & \cellred{0}      & \cellgreen{100} & \cellred{0} \\
    derive\_completion   & \cellgreen{100} & \cellgreen{100} & \cellorange{33}  & \cellred{0}     & \cellred{0} \\
    derive\_duration     & \cellgreen{100} & \cellgreen{100} & \cellyellow{67}  & \cellgreen{100} & \cellgreen{100} \\
    derive\_flags        & \cellgreen{100} & \cellgreen{100} & \cellyellow{67}  & \cellorange{33} & \cellgreen{100} \\
    derive\_demographics & \cellgreen{100} & \cellgreen{100} & \cellyellow{67}  & \cellgreen{100} & \cellgreen{100} \\
    derive\_site         & \cellgreen{100} & \cellgreen{100} & \cellyellow{67}  & \cellred{0}     & \cellgreen{100} \\
    derive\_baselines    & \cellgreen{100} & \cellgreen{100} & \cellgreen{100}  & \cellred{0}     & \cellred{0} \\
    derive\_study\_dates & \cellgreen{100} & \cellgreen{100} & \cellyellow{67}  & \cellred{0}     & \cellred{0} \\
    validate            & \cellgreen{100} & \cellgreen{100} & \cellgreen{100}  & \cellyellow{67} & \cellred{0} \\
    apply\_metadata     & \cellgreen{100} & \cellgreen{100} & \cellyellow{67}  & \cellred{0}     & \cellred{0} \\
    export              & \cellgreen{100} & \cellgreen{100} & \cellgreen{100}  & \cellred{0}     & \cellred{0} \\
    \midrule
    \textbf{Total}      & \textbf{100\%} & \textbf{100\%} & \textbf{73\%}   & \textbf{49\%}  & \textbf{53\%} \\
    \bottomrule
\end{tabular}

\vspace{0.2cm}
{\scriptsize
\cellcolor{green!30}\phantom{XX} 100\% \quad
\cellcolor{yellow!40}\phantom{XX} 67\% \quad
\cellcolor{orange!40}\phantom{XX} 33\% \quad
\cellcolor{red!30}\phantom{XX} 0\%
}

\caption{Per-node success rate heatmap (DAG v5, 15 nodes $\times$ 5 models). Values show success percentage. Claude models (Opus, Sonnet) achieve uniform 100\%. GPT-4.1 shows targeted failures on \texttt{derive\_disposition} (0\%) and \texttt{derive\_completion} (33\%). GPT-4o fails on intermediate derivations and downstream cascade nodes (49\% overall). Gemini fails on \texttt{validate} and all downstream nodes (53\%).}
\label{fig:heatmap}
\end{figure}

Three categories emerge:
\textbf{Universally easy} (100\%): \texttt{ingest}, \texttt{merge\_dm}---straightforward file loading.
\textbf{Model-discriminating}: \texttt{derive\_disposition} (GPT-4.1: 0/3) and \texttt{derive\_completion} (GPT-4o: 0/3) require specific dplyr idioms.
\textbf{Downstream-sensitive}: \texttt{validate}, \texttt{apply\_metadata}, \texttt{export} fail due to cascade effects.

\subsection{Generalizability: ADAE}

We evaluated on ADAE (adverse events), which differs from ADSL: 9 nodes with branching topology (merging ADSL and AE domains), BDS structure (one-record-per-event), 55 variables, 1{,}191 records.
Claude Sonnet achieves \textbf{100\% structural match} (55/55 variables, 1{,}191 records) in both runs.
All five L5 spot-checks pass (225 subjects, TRTEMFL=1{,}126, AESER=3, SAFFL=1{,}191).
Systematic L5 yields 67.6\% column accuracy (23/55 exact)---lower than ADSL due to MedDRA text-term formatting differences.
Wall-clock: 174s (vs.\ ADSL's 662s).

\subsection{DAG Design Iteration}

Figure~\ref{fig:progression} traces the progression from DAG v1 (9 nodes) to v5 (15 nodes).
Three changes drove 0\%$\to$100\%:


\begin{figure}[t]
\centering
\begin{tikzpicture}
\begin{axis}[
    width=0.85\textwidth,
    height=5.5cm,
    xlabel={DAG Version},
    ylabel={Structural Match (\%)},
    xmin=0.5, xmax=5.5,
    ymin=-5, ymax=110,
    xtick={1,2,3,4,5},
    xticklabels={v1, v2, v3, v4, v5},
    ytick={0, 20, 40, 59.2, 80, 100},
    yticklabels={0, 20, 40, 59.2, 80, 100},
    legend style={at={(0.02,0.98)}, anchor=north west, font=\scriptsize, fill=white, fill opacity=0.9, draw=gray!50},
    grid=major,
    grid style={gray!20},
    every axis plot/.append style={thick, mark size=3pt},
]

\addplot[color=blue, mark=*, solid]
    coordinates {(1,0) (2,42.9) (5,100)};
\addlegendentry{Claude Sonnet 4.6}

\addplot[color=orange, mark=square*, solid]
    coordinates {(1,0) (2,0) (3,0) (4,28.6) (5,59.2)};
\addlegendentry{GPT-4.1}

\addplot[color=green!60!black, dashed, mark=none, thick]
    coordinates {(0.5,59.2) (5.5,59.2)};
\addlegendentry{Keyword-RAG (Sonnet)}

\node[font=\tiny, anchor=south, blue] at (axis cs:2,42.9) {+schema};
\node[font=\tiny, anchor=south west, blue] at (axis cs:4.8,100) {+6 nodes};
\node[font=\tiny, anchor=north, orange] at (axis cs:4,28.6) {+hints};
\node[font=\tiny, anchor=south west, orange] at (axis cs:5,59.2) {+trt fix};

\end{axis}
\end{tikzpicture}
\caption{Structural match progression across DAG design iterations. Claude Sonnet improves from 0\% (v1) to 100\% (v5) through schema introspection (+42.9\,pp) and node expansion (+57.1\,pp); Sonnet v3--v4 were not run (hence the gap in the blue line). GPT-4.1 follows a slower trajectory, reaching 59.2\% at v5 after anti-hallucination hints. The dashed line shows the Keyword-RAG+Sonnet baseline (59.2\%), which DAG v5 surpasses for both models.}
\label{fig:progression}
\end{figure}
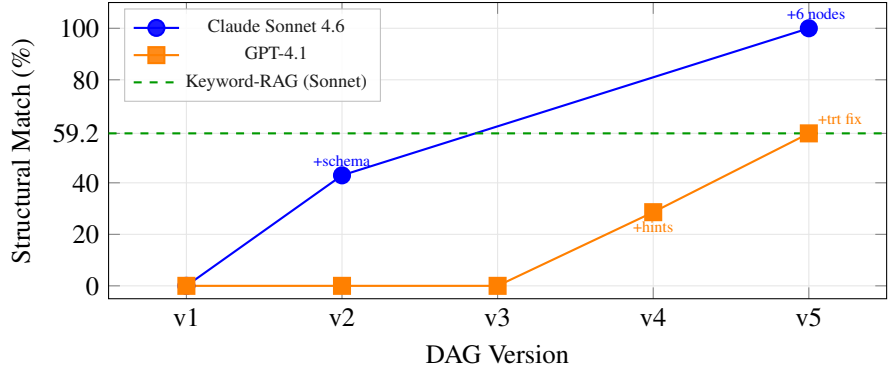

\textbf{(1) Schema introspection} (v1$\to$v2): Injecting upstream column names eliminated wrong-reference errors (Sonnet: 0\%$\to$42.9\%).
\textbf{(2) Node expansion} (v2$\to$v5): Adding 6 derivation nodes closed the coverage gap---the 9-node DAG structurally could not produce all 49 variables.
\textbf{(3) Anti-hallucination hints} (v4$\to$v5): Explicit ``use \texttt{as.Date()}, NOT admiral date functions'' fixed GPT-4.1's hallucinated \texttt{admiral::convert\_dtc\_to\_date()}, unblocking a 3-node cascade (+10.2\,pp).

\subsection{Statistical Summary}

DAG v5 node completion: 74.5\% (123/165; 95\% CI: 67.5\%--80.5\%, Wilson).
Claude: 100\% (60/60 nodes, 4 runs). GPT-4.1: 73.3\% (33/45, 3 runs). GPT-4o: 48.9\% (22/45, 3 runs).
L5 value spot-checks: all Claude runs produce placebo\_n=86, mean\_age=75.21, ITTFL\_Y=254, SAFFL\_Y=254, matching ground truth.

\textbf{L5 systematic evaluation.}
Column-by-column value comparison across all four Claude runs yields 73.7\% mean column accuracy (SD=1.2\%): 26/49 columns exact (Opus) and 72.1--74.9\% (Sonnet).
Identifiers and demographics (STUDYID, USUBJID, AGE, SEX, RACE, ARM) are universally correct; failures concentrate in numeric precision (BMIBL, HEIGHTBL---floating-point rounding) and complex derivation logic (COMP8FL, DCSREAS---multi-step conditionals).
The low cross-run variance indicates systematic errors (specification interpretation), suggesting targeted node-level improvements rather than more runs.

\section{Ablation Studies}
\label{sec:ablations}

We isolate five key design decisions. All ablations use CDISCPilot01 ADSL.

\textbf{A1: Architecture is decisive.}
Table~\ref{tab:main-results} shows architecture---not model capability---determines success.
Claude Sonnet scores 0\%/0\%/59.2\%/100\% under SingleAgent/FlatMulti/Keyword-RAG/DAG respectively; GPT-4.1 scores 0\%/0\%/0\%/59.2\%.

\textbf{A2: Topology alone is necessary but not sufficient.}
\emph{This isolates the contribution of DAG topology from node-specific prompt engineering.}
We test three levels of prompt specificity with the same 15-node ADSL DAG (Table~\ref{tab:topology-ablation}).
\textbf{Generic prompts} (step number, workspace listing, SDTM path only---no node-type hints, no skills, no spec text): 2 runs achieve 73\% mean node completion but 0\% structural match; the validate node merges incorrect intermediate artifacts, yielding 0 records.
\textbf{Topology + node-type hints} (adds INGEST/MERGE/DERIVE pattern guidance, node descriptions, skill context---but no specification text or expected output variables): 3 runs achieve 71\% mean node completion and 45.6\% mean structural match; 2/3 runs produce 254 records with correct value spot-checks.
\textbf{Full DAG} (adds specification text, expected output variables, anti-hallucination hints): 3 runs achieve 100\% node completion and 100\% structural match.
This gradient confirms that DAG topology provides the \emph{structural scaffold} (preventing catastrophic single-shot failure), node-type prompts provide \emph{derivation patterns} (lifting SM from 0\% to $\sim$46\%), and specification grounding provides \emph{variable-level precision} (lifting SM to 100\%).

\begin{table}[t]
    \centering
    \caption{Topology ablation (A2). Same 15-node DAG with progressive prompt specificity. Claude Sonnet 4.6, $n$ runs each.}
    \label{tab:topology-ablation}
    \begin{tabular}{@{}lrrrr@{}}
        \toprule
        Condition & Runs & Nodes & Struct\% & Records \\
        \midrule
        Topology only (generic) & 2 & 11/15 & 0.0 & 0 \\
        Topology + hints (no spec) & 3 & 10.7/15 & 45.6 & 2/3 \\
        Full DAG (all context) & 3 & 15/15 & 100.0 & 3/3 \\
        \bottomrule
    \end{tabular}
\end{table}

\textbf{A3: Schema introspection drives +42.9\,pp.}
\emph{This isolates the contribution of runtime schema context from other DAG components.}
DAG v1 (same 9-node topology, node-type prompts, retry) but \emph{without} schema introspection: Sonnet completes 8/9 nodes but produces 0\% structural match (wrong column references).
DAG v2 adds schema introspection---injecting upstream \texttt{.rds} column names into each worker prompt---yielding 42.9\% (21/49 vars) with the same topology and prompts.
GPT-4.1 shows a similar pattern: v1 completes 6/9 nodes (0\%), v2 completes 5/9 (0\%), indicating that schema context helps Sonnet more than GPT-4.1.
The +42.9\,pp gain from schema injection alone (topology held constant) confirms that runtime context is a necessary complement to topological decomposition.

\textbf{A4: Model tier structure.}
\emph{This tests whether model capability interacts with DAG architecture.}
Holding architecture constant (DAG v5), Table~\ref{tab:model-ablation} shows a clear three-tier ranking.
Claude models achieve 100\% node completion; GPT-4.1/4o succeed on simple nodes but fail on derivations requiring precise dplyr idioms; Gemini completes 53.3\% of nodes individually but fails on \texttt{validate}.

\begin{table}[t]
    \centering
    \caption{Model ablation on DAG v5 (A4). Node completion rate and structural match by worker model.}
    \label{tab:model-ablation}
    \begin{tabular}{@{}lrrrr@{}}
        \toprule
        Model & Runs & Nodes & Rate & Struct\% \\
        \midrule
        Claude Opus 4.6 & 1 & 15/15 & 100\% & 100.0 \\
        Claude Sonnet 4.6 & 3 & 45/45 & 100\% & 100.0 \\
        GPT-4.1 & 3 & 33/45 & 73.3\% & 59.2 \\
        GPT-4o & 3 & 22/45 & 48.9\% & 33.3 \\
        Gemini 2.5 Pro & 1 & 8/15 & 53.3\% & 0.0 \\
        \bottomrule
    \end{tabular}
\end{table}

\textbf{A5: Retry provides +9.7\,pp lift.}
\emph{This tests whether retry logic contributes beyond first-attempt success.}
Table~\ref{tab:retry-ablation} shows 107/165 nodes (64.8\%) succeed on first attempt; retry recovers 16 additional nodes (74.5\% total).
The effect is model-dependent: retry lifts Claude Sonnet from 86.7\% to 100\% (+13.3\,pp) and GPT-4.1 from 53.3\% to 73.3\% (+20.0\,pp), but provides no lift for GPT-4o.

\begin{table}[t]
    \centering
    \caption{Retry ablation (A5). First-attempt vs.\ with-retry node completion rates.}
    \label{tab:retry-ablation}
    \begin{tabular}{@{}lrrrrr@{}}
        \toprule
        Model & Nodes & Att.\,1 & +Retry & Fail & Lift \\
        \midrule
        Opus & 15 & 15 (100\%) & 0 & 0 & 0\,pp \\
        Sonnet & 45 & 39 (86.7\%) & 6 & 0 & +13.3\,pp \\
        GPT-4.1 & 45 & 24 (53.3\%) & 9 & 12 & +20.0\,pp \\
        GPT-4o & 45 & 22 (48.9\%) & 0 & 23 & 0\,pp \\
        Gemini & 15 & 7 (46.7\%) & 1 & 7 & +6.7\,pp \\
        \midrule
        \textbf{All} & \textbf{165} & \textbf{107 (64.8\%)} & \textbf{16} & \textbf{42} & \textbf{+9.7\,pp} \\
        \bottomrule
    \end{tabular}
\end{table}

\section{Discussion}
\label{sec:discussion}

\textbf{Why DAG topology works.}
Single-shot generation fails because ADSL requires coordinating 49 variables across 8 SDTM domains---exceeding current LLMs' reliable planning capacity.
The DAG addresses this via:
\textbf{(i)~Context reduction}---each node prompt contains $\sim$2K tokens (upstream schema + node spec) vs.\ $\sim$15K for the monolithic specification;
\textbf{(ii)~Error isolation}---failure at node $k$ preserves artifacts from nodes $1$--$(k{-}1)$, enabling retry without upstream loss;
\textbf{(iii)~Targeted prompts}---anti-hallucination hints are practical only at node granularity.
This decomposition benefits weaker models disproportionately: Claude produces longer, defensive code (56--308 lines/node with \texttt{tryCatch}, type coercion) achieving 100\%; GPT-4.1 (26--88 lines/node) reaches 59--73\% on dplyr-intensive nodes; GPT-4o scores 33--49\% with cascade failures.

\textbf{Why DAG beats retrieval baselines.}
Keyword-RAG+Sonnet achieves 59.2\% (29/49 vars)---the only non-DAG success.
Embedding-RAG+Sonnet retrieves 15 semantically similar pharmaverse docs via ChromaDB but still scores 0\%, confirming that contextual augmentation alone cannot decompose a 49-variable, 8-domain task.
DAG dominates on \emph{completeness} (49/49 vs.\ 29/49) and \emph{model robustness} (enables GPT-4.1 to match Keyword-RAG+Sonnet at 59.2\%, where Keyword-RAG+GPT-4.1 scores 0\%).

\textbf{Failure modes.}
When the DAG fails, it produces partial but inspectable results: each completed node's \texttt{.rds} artifact is preserved with per-node logs identifying the exact failure point.
Users can fix the failing node and resume---aligning with GxP audit trail requirements, unlike single-shot failure which produces no usable output.

\section{Related Work}

\textbf{LLM-based code generation.}
Benchmarks such as HumanEval~\citep{chen2021evaluating}, MBPP~\citep{austin2021program}, LiveCodeBench~\citep{jain2024livecodebench}, BigCodeBench~\citep{zhuo2024bigcodebench}, and SWE-bench~\citep{jimenez2024swebench} evaluate LLMs on general programming.
SWE-agent~\citep{yang2024sweagent} and self-debugging~\citep{chen2023selfdebug} improve code quality through agentic interfaces and iterative refinement.
Our work targets a specialized compliance domain with unique challenges not captured by general benchmarks.

\textbf{Multi-agent architectures.}
Multi-agent LLM systems have been explored for software development~\citep{wu2023autogen,qian2024chatdev,wang2024opendevin}, scientific discovery~\citep{hong2024metagpt}, and data analysis~\citep{zhang2023datacopilot}, combined with chain-of-thought~\citep{wei2022chain} and ReAct~\citep{yao2023react}.
MASAI~\citep{arora2024masai} and CodeR~\citep{chen2024coder} decompose software engineering via sub-agents and task graphs; Agentless~\citep{xia2024agentless} shows structured pipelines can match agentic approaches.
GxP-Agent differs in using \emph{predetermined, domain-knowledge-driven} decomposition: each DAG node maps to a concrete derivation step from the ADaM specification, providing variable-level coverage guarantees.

\textbf{Clinical NLP and domain-specific code generation.}
Prior clinical trial automation work has focused on protocol understanding and note processing~\citep{weng2024clinical}; ClinicalAgent~\citep{yue2024clinicalagent} targets outcome prediction, and Yang et al.~\citep{yang2024clinical_tfl} generate TLFs from ADaM data.
Domain-specific code generation for scientific computing~\citep{guo2024dsagent} and data science~\citep{zhang2023datacopilot} shows that domain context improves quality, but our results suggest that \textbf{structural decomposition} (the DAG) matters more than \textbf{contextual augmentation} (skills/retrieval).

\section{Limitations}
\label{sec:limitations}

\textbf{(1)} The topology-only ablation (A2) shows a 0\%$\to$100\% gradient but the full DAG combines multiple prompt choices not disentangled via factorial design.
\textbf{(2)} Stronger retrieval baselines might narrow the gap.
\textbf{(3)} All experiments use CDISCPilot01 (ADSL/ADAE); generalization to other studies or datasets is untested.
\textbf{(4)} Claude's 100\% vs.\ GPT-4.1's 59\% may partly reflect differential training data exposure.
\textbf{(5)} DAG requires 15 sequential LLM calls (662s vs.\ 65s for Keyword-RAG).
\textbf{(6)} With 1--3 runs per configuration, confidence intervals remain wide (74.5\%, 95\% CI: 67.5--80.5\%).
\textbf{(7)} The 15-node DAG was manually designed ($\sim$40 hours); automated generation is future work.
Additional limitations are in Appendix~\ref{app:additional-limitations}.

\section{Conclusion}

We presented GxP-Agent, a multi-agent system encoding regulatory process ordering as DAG topology for clinical trial programming.
On CDISC-Bench, GxP-Agent achieves 100\% structural match on ADSL with Claude (1 Opus + 3 Sonnet runs, 0\% variance)---a task where all single-shot approaches score 0\% and the best retrieval-augmented baseline achieves 59.2\%.
The core insight is that clinical trial programming is a compliance task: the correct decomposition and ordering of derivation steps can be determined from domain specifications \emph{a priori} and encoded as graph topology, reducing the LLM's role to generating correct R code for narrow, well-scoped subtasks.
We release CDISC-Bench and anonymized code as supplementary material to enable further research; full code and data will be released upon publication.

\bibliographystyle{plainnat}
\bibliography{references}

\appendix
\section{ADSL DAG Node Descriptions}
\label{app:nodes}

Table~\ref{tab:node-descriptions} provides detailed descriptions of each node in the 15-node ADSL process DAG, including the node type, input SDTM domains, and output variables.

\begin{table}[ht]
    \centering
    \caption{ADSL DAG v5: 15-node descriptions with types, source domains, and output variables.}
    \label{tab:node-descriptions}
    \footnotesize
    \resizebox{\textwidth}{!}{%
    \begin{tabular}{@{}rllll@{}}
        \toprule
        \# & Node & Type & Sources & Output Variables \\
        \midrule
        1 & ingest & INGEST & DM, EX, DS, MH, SC, VS, QS, SV & Load SDTM .xpt files into workspace \\
        2 & merge\_dm & MERGE & DM & STUDYID, USUBJID, SUBJID, SITEID, ARM, ARMCD, AGE, AGEU, SEX, RACE, ETHNIC \\
        3 & derive\_treatment\_vars & DERIVE & EX & TRT01P, TRT01A, TRT01PN, TRT01AN, TRTP, TRTA \\
        4 & derive\_trt\_dates & DERIVE\_DATE & EX & TRTSDT, TRTEDT \\
        5 & derive\_disposition & DERIVE & DS & EOSSTT, DCSREAS, DTHFL \\
        6 & derive\_completion & DERIVE & DS & COMP8FL, COMP16FL, COMP24FL \\
        7 & derive\_duration & DERIVE & (derived) & TRTDURD \\
        8 & derive\_flags & DERIVE & (derived) & SAFFL, ITTFL, EFFFL \\
        9 & derive\_demographics & DERIVE & DM, SC & AGEGR1, AGEGR1N, RACEN, HEIGHTBL, WEIGHTBL, BMIBL, BMIBLGR1, EDUCLVL \\
        10 & derive\_site & DERIVE & DM & SITEGR1 \\
        11 & derive\_baselines & DERIVE & VS, QS & HEIGHTBL, WEIGHTBL, BMIBL, BMIBLGR1, MMSETOT \\
        12 & derive\_study\_dates & DERIVE\_DATE & SV, DS & VISIT1DT, RFSTDTC, RFENDTC, DTHDT, DTHDTC, LSTALVDT \\
        13 & validate & VALIDATE & (all) & Verify: 1 rec/subj, 254 records, required vars present \\
        14 & apply\_metadata & METADATA & metacore & Apply variable labels, lengths, types from spec \\
        15 & export & EXPORT & (all) & Write final ADSL.xpt \\
        \bottomrule
    \end{tabular}%
    }
\end{table}

\section{CDISC-Bench Ground-Truth Variables}
\label{app:variables}

The ADSL ground truth contains 49 variables. Table~\ref{tab:gt-vars} lists all variable names organized by derivation category.

\begin{table}[ht]
    \centering
    \caption{CDISC-Bench ADSL ground-truth variable names (49 total).}
    \label{tab:gt-vars}
    \small
    \begin{tabular}{@{}lp{7.5cm}@{}}
        \toprule
        Category & Variables \\
        \midrule
        Identifiers (4) & STUDYID, USUBJID, SUBJID, SITEID \\
        Treatment (8) & ARM, ARMCD, TRT01P, TRT01A, TRT01PN, TRT01AN, TRTSDT, TRTEDT \\
        Demographics (8) & AGE, AGEU, AGEGR1, AGEGR1N, SEX, RACE, RACEN, ETHNIC \\
        Disposition (4) & EOSSTT, DCSREAS, DTHFL, DTHDT \\
        Completion (3) & COMP8FL, COMP16FL, COMP24FL \\
        Duration (1) & TRTDURD \\
        Flags (3) & SAFFL, ITTFL, EFFFL \\
        Baselines (5) & HEIGHTBL, WEIGHTBL, BMIBL, BMIBLGR1, MMSETOT \\
        Site (1) & SITEGR1 \\
        Education (1) & EDUCLVL \\
        Study dates (5) & VISIT1DT, RFSTDTC, RFENDTC, DTHDTC, LSTALVDT \\
        Other (6) & TRTP, TRTA, AVGDD, CUESSION, DISESSION, VIESSION \\
        \bottomrule
    \end{tabular}
\end{table}

\textbf{Note:} The ``Other'' category includes variables that appear in the CDISCPilot01 ground-truth XPT file. The structural match metric counts exact name matches between the generated output and this 49-variable list.

\section{Per-Node Success Rates}
\label{app:per-node}

Table~\ref{tab:per-node} reports per-node success rates across all DAG v5 runs.

\begin{table}[ht]
    \centering
    \caption{Per-node success rates across all DAG v5 runs (11 total: 1 Opus, 3 Sonnet, 3 GPT-4.1, 3 GPT-4o, 1 Gemini).}
    \label{tab:per-node}
    \small
    \begin{tabular}{@{}lcccccc@{}}
        \toprule
        Node & Opus & Sonnet & GPT-4.1 & GPT-4o & Gemini & Overall \\
        \midrule
        ingest & 1/1 & 3/3 & 3/3 & 3/3 & 1/1 & 11/11 \\
        merge\_dm & 1/1 & 3/3 & 3/3 & 3/3 & 1/1 & 11/11 \\
        derive\_treatment\_vars & 1/1 & 3/3 & 3/3 & 3/3 & 1/1 & 11/11 \\
        derive\_trt\_dates & 1/1 & 3/3 & 2/3 & 1/3 & 1/1 & 8/11 \\
        derive\_disposition & 1/1 & 3/3 & 0/3 & 3/3 & 0/1 & 7/11 \\
        derive\_completion & 1/1 & 3/3 & 1/3 & 0/3 & 0/1 & 5/11 \\
        derive\_duration & 1/1 & 3/3 & 2/3 & 3/3 & 1/1 & 10/11 \\
        derive\_flags & 1/1 & 3/3 & 2/3 & 1/3 & 1/1 & 8/11 \\
        derive\_demographics & 1/1 & 3/3 & 2/3 & 3/3 & 1/1 & 10/11 \\
        derive\_site & 1/1 & 3/3 & 2/3 & 0/3 & 1/1 & 7/11 \\
        derive\_baselines & 1/1 & 3/3 & 3/3 & 0/3 & 0/1 & 7/11 \\
        derive\_study\_dates & 1/1 & 3/3 & 2/3 & 0/3 & 0/1 & 6/11 \\
        validate & 1/1 & 3/3 & 3/3 & 2/3 & 0/1 & 9/11 \\
        apply\_metadata & 1/1 & 3/3 & 2/3 & 0/3 & 0/1 & 6/11 \\
        export & 1/1 & 3/3 & 3/3 & 0/3 & 0/1 & 7/11 \\
        \midrule
        \textbf{Total} & \textbf{15/15} & \textbf{45/45} & \textbf{33/45} & \textbf{22/45} & \textbf{8/15} & \textbf{123/165} \\
        \bottomrule
    \end{tabular}
\end{table}

\section{Extended DAG Version Changelog}
\label{app:versions}

\begin{itemize}[leftmargin=1.5em, itemsep=2pt]
    \item \textbf{v1} (9 nodes): Initial linear DAG. No schema injection. 8/9 nodes complete for Sonnet but validate fails (missing variables). Struct\% = 0\%.
    \item \textbf{v2} (9 nodes): Added schema introspection---inject upstream .rds column names into each worker prompt. Sonnet: 8/9 nodes, Struct\% = 42.9\% (21/49 vars). Key insight: schema context eliminates wrong-column-reference errors.
    \item \textbf{v3} (9 nodes): GPT-4.1 testing. 6/9 nodes, Struct\% = 0\%. Identified GPT-4.1-specific hallucination patterns (admiral function signatures).
    \item \textbf{v4} (9 nodes): Added anti-hallucination hints to node descriptions. GPT-4.1: 7/9 nodes, Struct\% = 28.6\%. Key fix: explicit ``do not use admiral::derive\_vars\_dt()'' guidance.
    \item \textbf{v5} (15 nodes): Expanded from 9 to 15 nodes. Added: derive\_treatment\_vars, derive\_demographics, derive\_baselines, derive\_completion, derive\_site, derive\_study\_dates. Added DERIVE\_DATE-specific hint for base R \texttt{as.Date()}. Sonnet: 15/15 (100\%), GPT-4.1: 9--12/15 (59.2\%).
\end{itemize}

\section{Additional Limitations}
\label{app:additional-limitations}

\textbf{(8) No human baseline.} We do not compare to qualified statistical programmer performance on the same task, precluding assessment of the practical gap between LLM-generated and human-written code.

\textbf{(9) L5 row-ordering caveat.} Our systematic value comparison assumes aligned row ordering between generated and ground-truth datasets; misaligned rows would undercount true matches. Cross-run L5 variance analysis is deferred.

\end{document}